%% file: main.tex
\documentclass{article}

\usepackage{microtype}
\usepackage{graphicx}
\usepackage{subfigure}
\usepackage{booktabs} %
\usepackage{enumitem}

\usepackage{hyperref}

\usepackage[dvipsnames]{xcolor}

\usepackage[accepted]{icml2024}

\usepackage{amsmath}
\usepackage{amssymb}
\usepackage{mathtools}
\usepackage{amsthm}

\usepackage[capitalize,noabbrev]{cleveref}

\renewcommand{\algorithmicrequire}{\textbf{Input:}}
\renewcommand{\algorithmicensure}{\textbf{Output:}}

\theoremstyle{plain}

\theoremstyle{definition}

\theoremstyle{remark}

\usepackage{tcolorbox}
\usepackage{xcolor,amsmath}

\DeclareMathOperator*{\argmin}{arg\,min}

\definecolor{codegreen}{rgb}{0,0.6,0}
\definecolor{codegray}{rgb}{0.5,0.5,0.5}
\definecolor{codepurple}{rgb}{0.58,0,0.82}
\definecolor{backcolour}{rgb}{0.97,0.97,0.97}
\usepackage{listings}
\lstdefinestyle{pythonstyle}{
    language=Python,
    backgroundcolor=\color{backcolour},   
    commentstyle=\color{codegreen},
    keywordstyle=\color{magenta},
    numberstyle=\ttfamily\tiny\color{codegray},
    stringstyle=\color{codepurple},
    basicstyle=\ttfamily\tiny,
    breakatwhitespace=false,         
    breaklines=true,                 
    captionpos=b,                    
    keepspaces=true,                 
    numbers=left,                    
    numbersep=5pt,                  
    showspaces=false,                
    showstringspaces=false,
    showtabs=false,                  
    tabsize=2,
    linewidth=0.95\textwidth,
    xleftmargin=0.05\textwidth,
}
\lstdefinestyle{markdownstyle}{
    basicstyle=\ttfamily\tiny,
    backgroundcolor=\color{backcolour},   
    xleftmargin=0.05\textwidth,
    xrightmargin=0.05\textwidth,
    breakindent=0\dimen0,
    columns=flexible,
    showspaces=false,
    showstringspaces=false,
    breaklines=true,
    breakatwhitespace=true,
    breakautoindent=true,
}

\newcommand{\METHOD}{Scientific Generative Agent}
\newcommand{\ACRONYM}{SGA}

\renewcommand{\eqref}{Eq.~\ref}
\newcommand{\secref}{Sec.~\ref}
\newcommand{\figref}{Fig.~\ref}

\newcommand{\algref}{Alg.~\ref}

\usepackage[textsize=tiny]{todonotes}

\usepackage{wrapfig}
\usepackage{multirow}
\usepackage{amssymb}
\usepackage{pifont}
\newcommand{\cmark}{\text{\ding{51}}}
\newcommand{\xmark}{\text{\ding{55}}}
 
\icmltitlerunning{\METHOD{}}

\begin{document}

\twocolumn[

\icmltitle{LLM and Simulation as Bilevel Optimizers:\\
A New Paradigm to Advance Physical Scientific Discovery}

\icmlsetsymbol{equal}{*}

\begin{icmlauthorlist}
\icmlauthor{Pingchuan Ma}{csail}
\icmlauthor{Tsun-Hsuan Wang}{csail}
\icmlauthor{Minghao Guo}{csail}
\icmlauthor{Zhiqing Sun}{cmu}
\\
\icmlauthor{Joshua B. Tenenbaum}{csail,bcs,cbmm}
\icmlauthor{Daniela Rus}{csail}
\icmlauthor{Chuang Gan}{umass,ibm}
\icmlauthor{Wojciech Matusik}{csail}
\end{icmlauthorlist}

\icmlaffiliation{csail}{MIT CSAIL}
\icmlaffiliation{cmu}{CMU LTI}
\icmlaffiliation{bcs}{MIT BCS}
\icmlaffiliation{cbmm}{Center for Brains, Minds and Machines}
\icmlaffiliation{umass}{UMass Amherst}
\icmlaffiliation{ibm}{MIT-IBM Watson AI Lab}

\icmlcorrespondingauthor{Pingchuan Ma}{pcma@csail.mit.edu}

\icmlkeywords{AI4Science, Large Language Model, Language Agent, Physical Simulation}

\vskip 0.3in
]

\printAffiliationsAndNotice{}  %

\input{src/0_abstract}
\input{src/1_introduction}

\input{src/3_method}

\input{src/4_experiments}
\input{src/2_related_work}
\input{src/5_conclusion}
\input{src/6_impact}

\bibliography{main}
\bibliographystyle{icml2024}

\newpage
\input{src/n_appendix}

\end{document}

%% file: src/0_abstract.tex
\begin{abstract}
Large Language Models have recently gained significant attention in scientific discovery for their extensive knowledge and advanced reasoning capabilities. However, they encounter challenges in effectively simulating observational feedback and grounding it with language to propel advancements in physical scientific discovery. Conversely, human scientists undertake scientific discovery by formulating hypotheses, conducting experiments, and revising theories through observational analysis. Inspired by this, we propose to enhance the knowledge-driven, abstract reasoning abilities of LLMs with the computational strength of simulations. We introduce \METHOD{} (\ACRONYM{}), a bilevel optimization framework: LLMs act as knowledgeable and versatile thinkers, proposing scientific hypotheses and reason about discrete components, such as physics equations or molecule structures; meanwhile, simulations function as experimental platforms, providing observational feedback and optimizing via differentiability for continuous parts, such as physical parameters. We conduct extensive experiments to demonstrate our framework's efficacy in constitutive law discovery and molecular design, unveiling novel solutions that differ from conventional human expectations yet remain coherent upon analysis.
\end{abstract}

%% file: src/1_introduction.tex
\section{Introduction}
\label{sec:introduction}

In physical science, spanning physics, chemistry, pharmacology, etc., various research streams aim to automate and speed up scientific discovery \citep{wang2023scientific}. Each stream innovates within its field, creating methods tailored to its specific challenges and nuances. However, this approach often misses a universally applicable philosophy \citep{popper2005logic,fortunato2018science}, which can be pivotal to democratizing access to advanced research tools, standardizing scientific practices, and enhancing efficiency across disciplines. Our goal aims to transcend specific domains, offering a unified approach to physical science.

As an inspiration, we observe how human scientists conduct scientific discovery experiments and conclude a few key experiences: (i) iteratively propose a hypothesis and make observations from experimentation to correct theories \citep{popper2005logic}, (ii) divide the solutions into discrete components, such as physics equations or molecule structures, and continuous components, such as parameters for physics and molecule properties \citep{wang2023scientific}, (iii) exploit the existing knowledge while occasionally explore novel ideas aggressively in pursuits of breakthrough \citep{wuestman2020typology}, (iv) follow a generic, universal principle for all types of physical scientific discovery yet with specific nuance of each discipline \citep{rosenberg2019philosophy}.

\begin{figure*}[t]
\centering
\includegraphics[width=\linewidth]{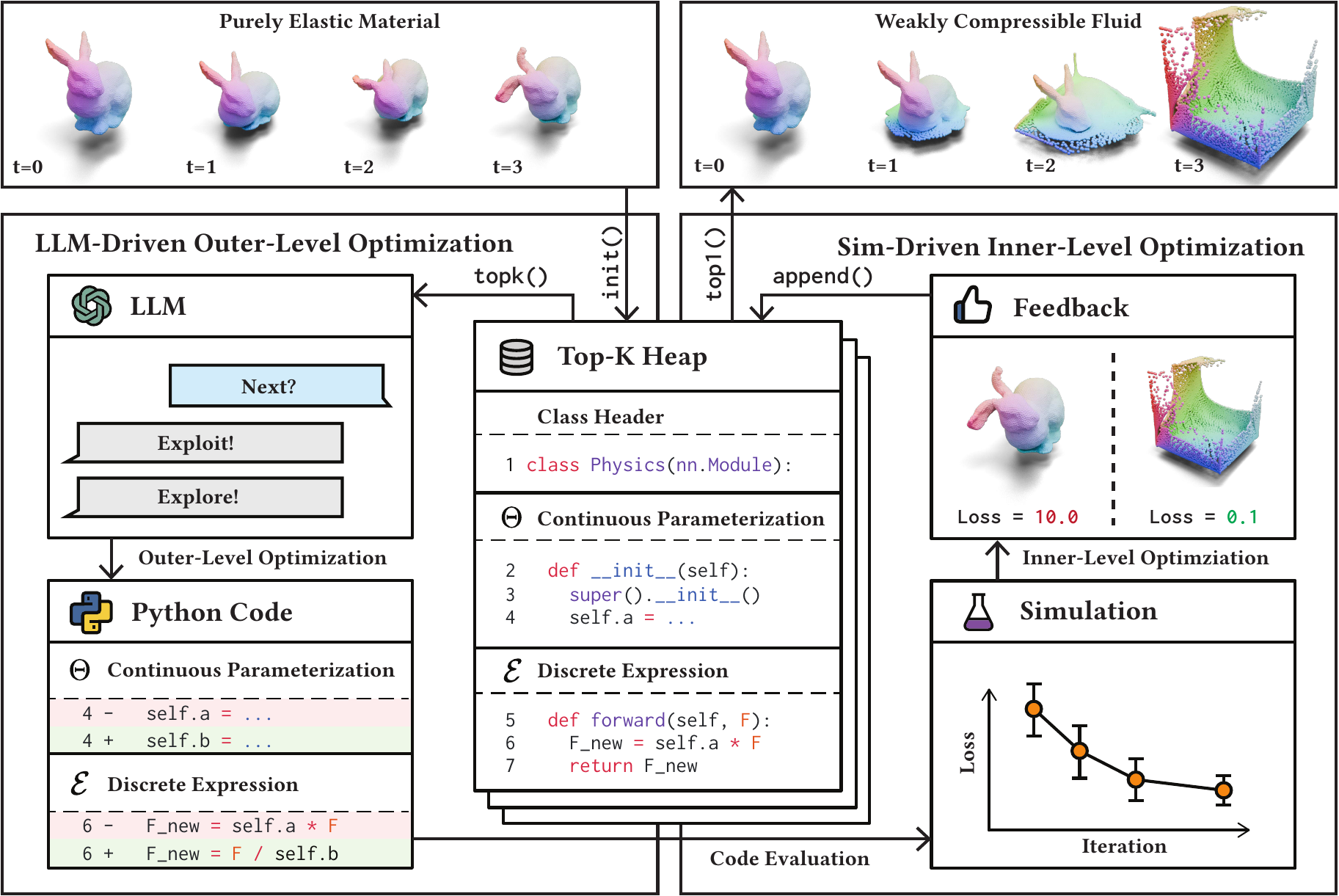}
\vspace{-5mm}
\caption{\textbf{The overall pipeline of \METHOD{} (\ACRONYM{}).} Taking the constitutive law searching problem as an example, the input is an initial guess (a purely elastic material), and the output is another constitutive law optimized towards the ground-truth (weakly compressible fluid). The initial guess first initialize a top-$K$ heap for storing the solutions. In the outer-level optimization, an LLM takes in top-$K$ previously proposed solutions and generates a better one upon them with modified continuous parameterization $\Theta$ and discrete expression $\mathcal{E}$. In the inner-level optimization, a gradient-based optimization solves for optimal $\Theta$ via simulation and appends these optimized solutions in the heap. After a few iterations of bilevel optimization, the heap returns the top-1 solutions as the final solution.}
\vspace{-2mm}
\label{fig:pipeline}
\end{figure*}

Standing out as generalist tools with an extensive repository of knowledge \citep{ai4science2023impact}, large language models (LLMs) have recently risen to prominence in scientific discovery for their expansive knowledge bases, advanced reasoning capabilities, and human-friendly natural language interface. One line of research focuses on fine-tuning LLMs with domain-specific data to align natural language with scientific information, such as chemical \citep{chithrananda2020chemberta} or drug \citep{liu2021ai} structures; however, these methods are domain-bound and demand extensive data for broader application. Another research direction seeks to leverage the innate capabilities of pre-trained LLMs, augmented by external resources like the internet, programming, or documentation. LLMs serve as optimizers or agents \citep{huang2023benchmarking} for mathematical problem-solving \citep{romera2023mathematical}, conducting chemical experiments \citep{boiko2023autonomous}, and advancing molecular \citep{li2023empowering} and drug discovery \citep{sharma2023chatgpt}. Nevertheless, these approaches are confined to the computational capability of LLMs, a crucial factor in physical science for tasks like calculating numerical results based on physics law hypotheses to predict natural phenomena. To address this limitation, we propose to augment LLMs with physical simulation, hereby merging the knowledge-driven, abstract reasoning abilities of LLMs with the computational structures and accuracy of simulations.

To this end, inspired by the overarching philosophy of human scientists, 
we introduce \textbf{\METHOD{} (\ACRONYM{})}, a bilevel optimization approach wherein the outer-level engages LLMs as knowledgeable and versatile thinkers for generating and revising scientific hypothesis, while the inner-level involves simulations as experimental platforms for providing observational feedback.
First, we employ LLMs to generate hypotheses, which then guide the execution of simulations. These simulations, in turn, yield observational feedback that helps refine and improve the proposed hypotheses. Secondly, we introduce a bilevel optimization framework: one level performs search-based optimization on discrete, symbolic variables like physics laws or molecule structures via LLMs; the other level performs gradient-based optimization via differentiable simulation for continuous parameters like material stiffness or molecule coordinates. Thirdly, we devise an exploit-and-explore strategy for the hypothesis proposal by adjusting LLM's generation temperature. Lastly, we demonstrate our pipeline is generally applicable across scientific disciplines, with only minimal modification such as altering the prompts.

For the empirical study, we focus on (i) \textit{molecular design} that aims to discover molecular structure and atoms' coordinates based on its conformation and quantum mechanical properties and (ii) \textit{constitutive law discovery} that aims to discover material constitutive equations and its corresponding mechanical properties directly from a recorded motion trajectory. To provide a concrete example, let's assume that we initially have simply the code for a purely linear material. We then task our model to uncover a more complex representation by optimizing its code to fit a highly non-linear trajectory. In this task, our method capitalizes on the strengths of bilevel optimization: the outer-level utilizes LLMs to identify the correct symbolic material constitutive equations and formulates a proposition for potentially beneficial continuous parameterization (e.g., Young's modulus and Poisson's ratio); and the inner-level refines the proposed material parameters and provides informative feedback using differentiable simulation. Generally, our method can discover the desired molecules and constitutive laws, outperforming other LLM-based baselines; more interestingly, it can propose well-performing solutions that are beyond human expectation yet sensible under analysis by domain experts.
Overall, our contributions are concluded as:
\begin{itemize}[align=right,itemindent=0em,labelsep=2pt,labelwidth=1em,leftmargin=*,itemsep=0em] 
    \item We present a generic framework for physical scientific discovery that combines LLMs with physical simulations.
    \item We propose a bilevel optimization with LLMs for discrete-space search-based optimization and differentiable simulations for continuous-space gradient-based optimization.
    \item We conduct extensive experiments to demonstrate the effectiveness and generality of the proposed framework in physics law discovery and molecular design; moreover, we showcase novel molecules or constitutive laws, while unexpected from a conventional perspective, are deemed reasonable upon examination by domain experts.
\end{itemize}

%% file: src/3_method.tex
\section{\METHOD{}}
\label{sec:method}

\ACRONYM{} is a bilevel optimization framework where the upper level features LLMs as proposers of scientific solutions, and the lower level utilizes simulations as experimental platforms for validation. In \secref{ssec:bilevel}, we describe a formal definition of the bilevel optimization, followed by \secref{ssec:outer_optim} for outer optimization and \secref{ssec:inner_optim} for inner optimization.

\subsection{Bilevel Optimization Pipeline}
\label{ssec:bilevel}

We formally describe the pipeline of our method, including the input/output of the system and the underlying submodules, and the overall optimization formulation. Suppose we are given a metric to evaluate a physical phenomenon $y$ (e.g., a configuration of deformation) for a scientific problem $\mathcal{L}\left(y\right)$ (e.g., reconstruction of a mechanistic behavior). First, we describe the simulation (as an experimental platform) as,
\begin{align}
    \label{eq:sim}
    y, z = \Phi\left(\theta ; \mathcal{E}\right),
\end{align}
where $\Phi$ is a simulator that takes in scientific expression $\mathcal{E}$ (e.g., constitutive equations) and continuous components $\theta$ (e.g., material parameters) as inputs and gives simulated physical phenomenon $y$ and additional observational feedback $z$ (e.g., particles' trajectories) as outputs. Next, the LLM is prompted to act as a thinker to propose expressions $\mathcal{E}$ based on past experimental results from simulation,
\begin{align}
    \label{eq:llm}
    \mathcal{E},\Theta = \text{LLM}\left(\left\{\mathcal{L}\left(y_k\right),z_k,o_k,\mathcal{E}_k,\Theta_k\right\}_{k\in[K]};\mathcal{P}\right),
\end{align}
where the set $[K]$ summarizes the pointers to the past simulation results containing an evaluation of the scientific problem $\mathcal{L}(y_k)$, other physical feedback $(z_k,o_k)$, and past proposals $(\mathcal{E}_k,\Theta_k)$; $o_k$ summarizes the intermediate results of the inner optimization (later detailed in \secref{ssec:inner_optim}); $\Theta$ determines the continuous parameterization for the decision variables of the inner optimization (e.g., which variables to be optimized within a proposed equation); $\mathcal{P}$ is prompt. With these, we define the bilevel optimization problem as,
\begin{subequations}
\label{eq:bilevel}
\begin{align}
    \underset{\mathcal{E},\Theta}{\min}~&\mathcal{L}\left(y\left(\mathcal{E},\Theta,\hat{\theta};\Phi\right)\right) \label{eq:bilevel_1} \\ 
    \text{s.t. }& G\left(\mathcal{E},\Theta; \Phi\right)\leq 0 \label{eq:bilevel_2} \\
    &\hat{\theta}\in \underset{\theta \in \Theta}{\argmin}~\mathcal{L}\left(y\left(\theta; \Phi, \mathcal{E}\right)\right), \label{eq:bilevel_3}
\end{align}
\end{subequations}
where $G\left(\cdot\right)\leq 0$ refers to the validity of the simulation (i.e., whether an expression $\mathcal{E}$ is simulatable). The outer optimization searches for (i) an expression $\mathcal{E}$ that defines what experiments to be conducted $\Phi\left(\cdot; \mathcal{E}\right)$ and (ii) continuous parametrization $\Theta$ that defines the search space of the inner continuous optimization $\min_{\theta \in \Theta}$. With the dependencies on the outer-level variables $(\mathcal{E},\Theta)$, the inner optimization searches for the optimal continuous parameters $\hat{\theta}$ given the proposed expression via differentiable simulation.

\begin{algorithm}[t]
\small
\caption{\METHOD{}}
\label{alg:sga}
\begin{algorithmic}[1]
\REQUIRE Discrete expression and continuous param \texttt{($\mathcal{E}$,$\theta\in\Theta$)},\\
~~~~~~Num of exploiting $M_l$, Num of exploring $M_h$,\\
~~~~~~Exploiting temperature $T_l$, Exploring temperature $T_h$
\STATE \texttt{\color{Green}\# Store ranked (solution,param) by heap}
\STATE \texttt{H $\leftarrow$ heap()}
\STATE \texttt{\color{Green}\# Continuous optimization}
\STATE \texttt{$\hat{\theta}$ $\leftarrow$ optim($\mathcal{E}$,$\theta$;$\Phi$)}
\STATE \texttt{H.append(($\mathcal{E}$,$\hat{\theta}$))}
\FOR {$i=1,\ldots,N$}
    \STATE \texttt{\color{Green}\# Generate $M_l$ solutions from LLM}
    \STATE \texttt{($\mathcal{E}$,$\Theta$)[:$M_l$] $\leftarrow$ LLM(H.topk($K$),$T_l$)}
    \STATE \texttt{\color{Green}\# Generate $M_h$ solutions from LLM}
    \STATE \texttt{($\mathcal{E}$,$\Theta$)[$M_l$:$M_l$+$M_h$] $\leftarrow$ LLM(H.topk($K$),$T_h$)}
    \FOR {$m=1,\ldots,M_l+M_h$}
        \STATE \texttt{\color{Green}\# Continuous optimization}
        \STATE \texttt{$\hat{\theta}$ $\leftarrow$ optim($\mathcal{E}$,$\theta\in\Theta$;$\Phi$)}
        \STATE \texttt{H.append(($\mathcal{E}$,$\hat{\theta}$)})
    \ENDFOR
\ENDFOR
\ENSURE \texttt{H.topk(1}) \texttt{\color{Green}\# Return the best}
\end{algorithmic}
\end{algorithm}

\subsection{LLM-Driven Outer-Level Search}
\label{ssec:outer_optim}

We dive deeper into how we use LLMs (\eqref{eq:llm}) and their interaction with the simulation for outer-level search (\eqref{eq:bilevel}).

\paragraph{LLM-driven Optimization}
LLMs have shown to be effective sequential decision makers for generic optimization, providing proper guidance via prompting and sufficiently informative contexts \citep{yang2024large,romera2023mathematical}. We craft prompts to direct LLMs in a structured manner, enabling them to (i) perform \textit{analysis} on past experimental results from the simulation, e.g., the deviatoric parts of stress tensor are likely correct based on the loss curve; (ii) devise a high-level \textit{plan} on how to formulate a hypothesis or improve upon previous experiments, e.g., ensure numerical stability with the usage of the determinant of deformation gradient; (iii) suggest a \textit{solution} that can be executed as experiments via simulation for hypothesis testing; e.g., a code snippet describing a constitutive equation. For \eqref{eq:bilevel_1}, inspired by \citep{ma2024eureka}, we adopt an evolutionary search that generates multiple offspring $\{\mathcal{E}_m,\Theta_m\}_{m\in[M]}$ ($M$ is offspring size) in each iteration and retain the best selection. Distinctively, our approach (\algref{alg:sga}) involves selecting several high-performing candidates rather than the best only, which (i) enhances the feasibility of hypotheses in simulation (\eqref{eq:bilevel_2}) and (ii) facilitates evolutionary crossover, with LLMs generating new hypotheses from various past experiments (``breeds'') for better exploration, akin to the findings in \citep{romera2023mathematical}.

\paragraph{Interfacing with Simulation}
The primary challenge in integrating LLMs with simulation lies in devising a protocol that enables efficient, structured, yet adaptable communication between the two modules. We observe that physical scientific solutions are often represented as mathematical expressions or structured entities. Hereby, from LLMs to simulation, we consider two settings: equation searching and entity searching, both unified as the abstraction $(\mathcal{E},\Theta)$ in \eqref{eq:llm}. In \textit{equation searching}, LLMs are allowed to propose equations $\mathcal{E}$ along with the search space of the inner-level continuous optimization $\Theta$; for practitioners, an example using PyTorch can be $\Theta$ as \texttt{\_\_init\_\_} that defines continuous parameters via \texttt{nn.Parameter} and $\mathcal{E}$ as \texttt{forward} that defines computation of equations (see \figref{fig:pipeline}). In \textit{entity searching}, LLMs propose descriptions of structures $\mathcal{E}$ (e.g., how atoms are connected to form a molecule) with $\Theta$ simply reduced to constant (e.g., every atom has its 3D coordinates to be optimized) and omitted from the optimization \eqref{eq:bilevel_1} as decision variables. On the other hand, from simulation to LLMs, we leverage domain experts' knowledge to craft functions for extracting compact, relevant information $z$ as observational feedback; this process is akin to an experienced scientist offering guidance to a junior colleague on how to document experimental findings effectively. For instance, human experts often monitor the movements of specific body regions to derive constitutive laws. Therefore, to aid in this process, we include a function in the simulation that records the particle trajectories. Lastly, the subsequent section \secref{ssec:inner_optim} will provide an in-depth explanation of the inner optimization results denoted as $o$. These results serve as feedback from the simulation to the LLMs.

\paragraph{Exploitation and Exploration}
Inspired by human scientists achieving breakthroughs by skillfully balancing careful progression with bold exploration, we devise an exploit-and-explore strategy by tuning the LLMs' decoding temperature \citep{yang2024large}. When generating offspring $\{\mathcal{E}_m,\Theta_m\}_{m\in[M]}$ in \eqref{eq:bilevel_1}, we divide them into two groups: one (${m\in\mathcal{M}_\text{exploit}}$) consists of cautious followers that keep the ``gradient'' and conservatively trails previous solutions, while the other (${m\in\mathcal{M}_\text{explore}}$) comprises daring adventurers that take risks and suggest unique solutions. Empirically, we observed that (i) $\mathcal{M}_\text{exploit}$ often contains repetitive solutions from previous iterations, and (ii) $\mathcal{M}_\text{explore}$ tends to yield solutions too random to be informative for guiding optimization, or invalid (i.e., violating \eqref{eq:bilevel_2}), thus providing little feedback signal. As a rule of thumb, we have found that a 1:3 ratio between $\mathcal{M}_\text{exploit}$ and $\mathcal{M}_\text{explore}$ is effective.

\subsection{Differentiable Inner-Level Optimization}
\label{ssec:inner_optim}
Under the search space $\Theta$ and expression for simulation $\mathcal{E}$ from the outer level, inner optimization (\eqref{eq:bilevel_3}) involves a gradient-based optimization that solves for optimal continuous parameters $\hat{\theta}\in\Theta$ via differentiable simulation 
(\eqref{eq:sim}). Essentially, the domain-specific knowledge is distilled via gradients $\nabla_\theta \Phi(\theta; \mathcal{E})$ from the simulation to the intermediate optimization results $o$ (like loss curve). The $(\hat{y},o)$ are then fed back to LLMs for revising solutions. Note that $o$ may involve the loss curve toward the target metric $\mathcal{L}$ and other auxiliary recordings throughout optimization, carrying information of how to improve solutions in various aspects; for example, with $\mathcal{L}$ as displacement of position, $o$ may include velocities across the inner optimization iterations.

%% file: src/4_experiments.tex
\begin{table*}[t]
\caption{\textbf{Benchmark}. We compare our method against 4 baselines and 2 variations of our method, while also noting the difference in architecture or hyper-parameters. We use column \textbf{\#Iter.} as the number of iterations, \textbf{\#Hist.} as the $K$ value for the top-k retrieval in the historical optimization steps, $\frac{\textbf{\#Exploit}}{\textbf{\#Explore}}$ as the number of offspring for exploitation versus exploration, \textbf{Bilevel} as if bilevel optimization is enabled. Our experiments encompass 8 different tasks, which are divided into constitutive law search \textbf{(a-d)} and molecule design \textbf{(e-h)}. A lower loss value is preferable across all tasks. The best method with the lowest loss is highlighted in \textbf{bold} text.}
\vspace{2mm}
\resizebox{\linewidth}{!}{
\begin{tabular}{l|cccc|cccc|cccc}
\toprule
\multirow{2}{*}{\textbf{Method}} & \multirow{2}{*}{\textbf{\#Iter.}} & \multirow{2}{*}{\textbf{\#Hist.}}  & \multirow{2}{*}{\textbf{$\frac{\textbf{\#Exploit}}{\textbf{\#Explore}}$}} & \multirow{2}{*}{\textbf{Bilevel}} & \multicolumn{4}{c|}{\textbf{Constitutive Law Search}} & \multicolumn{4}{c}{\textbf{Molecule Design}} \\ \cmidrule{6-13} 
 &  &  &  &  & \textbf{(a)} $\downarrow$ & \textbf{(b)} $\downarrow$ & \textbf{(c)} $\downarrow$ & \textbf{(d)} $\downarrow$ & \textbf{(e)} $\downarrow$ & \textbf{(f)} $\downarrow$ & \textbf{(g)} $\downarrow$ & \textbf{(h)} $\downarrow$
\\ \midrule
\textbf{CoT} & 1 & 5 & N/A & $\xmark$ &
298.5 & 1462.3 & 150.0 & 384.1 &
3.0 & 32.1 & 18.6 & 6.0
\\
\textbf{FunSearch} & 20 & 2 & 0 / 4 & $\xmark$ &
210.3 & 872.2 & 82.8 & 139.5 &
1.1 & 7.1 & 8.3 & 1.1
\\
\textbf{Eureka} & 5 & 1 & 0 / 16 & $\xmark$ &
128.0 & 531.0 & 101.7 & 150.1 &
4.3 & 9.8 & 3.3 & 9.7e-1
\\
\textbf{OPRO} & 5 & 5 & 0 / 16 & $\xmark$ &
136.2 & 508.3 & 99.2 & 128.8 &
2.4 & 9.4 & 3.1 & 1.3
\\ \midrule
\textbf{Ours (no bilevel)} & 5 & 5 & 4 / 12 & $\xmark$ &
90.2 & 517.0  & 83.6  & 68.4 &
8.6e-1 & 9.1 & 1.8 & 1.4
\\
\textbf{Ours (no exploit)} & 5 & 5 & 0 / 16 & $\cmark$ &
3.0e-3  & 3.9e-1 & 6.6e-2 & \textbf{1.4e-12} &
4.0e-4 & 1.5e-1 & 6.1e-1 & \textbf{2.8e-5}
\\ \midrule
\textbf{Ours} & 5 & 5 & 4 / 12 & $\cmark$ &
\textbf{5.2e-5} & \textbf{2.1e-1} & \textbf{6.0e-2} & \textbf{1.4e-12} &
\textbf{1.3e-4} & \textbf{1.1e-1} & \textbf{5.4e-1} & 3.6e-5
\\ \bottomrule
\end{tabular}
}
\vspace{-2mm}
\label{tbl:benchmark}
\end{table*}

\section{Experiments}
\label{sec:experiments}

\subsection{Problem Definitions}

\paragraph{Constitutive Law Discovery} Identifying the constitutive law from motion observations stands as one of the most difficult challenges in fields such as physics, material science, and mechanical engineering.
Here we follow the recent advances in physical simulation and formulate the constitutive law discovery task as an optimization problem~\cite{ma2023learning} using differentiable Material Point Method (MPM) simulators~\cite{sulsky1995application,jiang2016material}. 
Note that our method is not specifically tailored to MPM simulators and applies to any physical simulation.
The objective of this task is to identify both the discrete expression and continuous parameters in a constitutive law, specifically the symbolic material models $\varphi\left(\cdot\right)$ and their corresponding material parameters $\theta$, from a ground-truth trajectory of particle positions $\hat{X}_{t\in\left[1,\dots,T\right]}$ where $T$ denotes the number of steps. In this problem, we consider two types of constitutive laws, $\varphi_E\left(\cdot;\theta_E\right)$ and $\varphi_P\left(\cdot;\theta_P\right)$, for modeling elastic and plastic materials respectively, and they are formally defined as:
\begin{subequations}
\begin{align}
    \varphi_E\left(\mathbf{F};\theta_E\right)&\mapsto\boldsymbol{\tau}\\
    \varphi_P\left(\mathbf{F};\theta_P\right)&\mapsto\mathbf{F}^\text{corrected},
\end{align}
\end{subequations}
where $\mathbf{F}\in\mathbb{R}^{3\times3}$ is the deformation gradient, $\boldsymbol{\tau}\in\mathbb{R}^{3\times3}$ is the Kirchhoff stress tensor, $\mathbf{F}^\text{corrected}\in\mathbb{R}^{3\times3}$ is the deformation gradient after plastic return-mapping correction, and $\theta_E$ and $\theta_P$ are the continuous material parameters for elastic and plastic constitutive laws respectively. Given a specific constitutive law, we input it to the differentiable simulation and yields a particle position trajectory:
\begin{align}
    X_{t\in\left[1,\dots,T\right]}=\text{sim}\left(\varphi\left(\cdot;\theta\right)\right),
\end{align}
and we optimize the constitutive law by fitting the output trajectory to the ground truth $\hat{X}_{t\in\left[1,\dots, T\right]}$.

\paragraph{Molecule Design}
In this study, we focus on a prevalent task in molecule design: discovering molecules with specific quantum mechanical properties. 
Our objective is to determine the optimal molecular structure and its 3D conformation to match a predefined target quantum mechanical property. 
The design process involves both the discrete expression -- the molecular structure represented by SMILES strings~\citep{weininger1988smiles}, and the continuous parameters -- the 3D coordinates of each atom in the molecule.
The methodology comprises two loops: In the outer loop, the LLM generates the initial molecular structure as a SMILES string, along with a preliminary guess for the 3D atom coordinates. 
The inner loop involves simultaneous optimization of both the molecule's 3D conformation and quantum mechanical properties, both determined by 3D atom positions.

\begin{figure}[t]
    \centering
    \includegraphics[width=\linewidth]{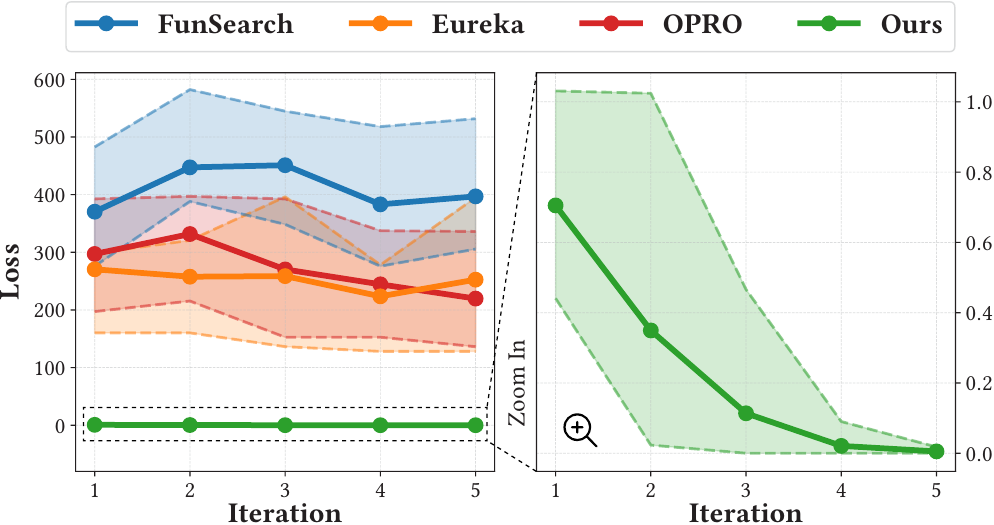}
    \vspace{-5mm}
    \caption{\textbf{Loss trends comparison.} Loss of the best solution averaged across seeds at different iterations of LLM-driven optimization, where the shading shows the min/max value.}
    \label{fig:benchmark}
    \vspace{-2mm}
\end{figure}

For the generation of 3D conformations, we utilize the ETKGD algorithm~\citep{riniker2015better} followed by optimization using the Merck Molecular Force Field (MMFF)~\citep{halgren1996merck}, both implemented within the RDKit~\citep{landrum2013rdkit}. To get the quantum mechanical property values, we employ UniMol~\citep{zhou2023uni}, a pre-trained transformer-based large model, which has been fine-tuned on the QM9 dataset~\citep{ramakrishnan2014quantum}.

\subsection{Experiment Setup}

\paragraph{Task Design} We design a diverse set of challenging tasks for evaluation. For constitutive law discovery, we propose 4 tasks including:
\textbf{(a)}~fitting the non-linear elastic material starting from a linear elastic material,
\textbf{(b)}~fitting the von Mises plastic material starting from a purely elastic material,
\textbf{(c)}~fitting the granular material starting from a purely elastic material,
and \textbf{(d)}~fitting the weakly compressible fluid starting from a purely elastic material.
For molecular design task, we consider 4 popular tasks, centering on 3 commonly evaluated quantum mechanical properties~\citep{fang2022geometry, zhou2023uni}, each set to different target values:
\textbf{(e)}~HOMO (Highest Occupied Molecular Orbital) set to 0,
\textbf{(f)}~LUMO (Lowest Unoccupied Molecular Orbital) set to 0,
\textbf{(g)}~the HOMO-LUMO energy gap set to 0,
and \textbf{(h)}~the HOMO-LUMO energy gap set to -2. 
All these values are normalized on all data in QM9 dataset. 

\paragraph{Implementation Details} We run all our experiments 5 times with different random seeds following previous practices~\citep{ma2024eureka}. Due to the complexity of the task, we provide a simple bootstrapping example of a valid design to ensure the success rate. We use warp~\citep{warp2022} for the differentiable MPM simulation, and we develop our inner-level optimization upon PyTorch \citep{paszke2019pytorch}. In all our experiments, we use mean square error as the criteria and Adam optimizer~\citep{KingBa15}. We choose \texttt{gpt-4-turbo-preview} as the backbone model for LLM and tentatively set the exploiting temperature $T_l=0.5$ and exploring temperature $T_h=1.0$.

\begin{table}[t]
\centering
\caption{\textbf{Comparison with symbolic regression}. We compare our method against 5 most performant methods in SRBench~\cite{cava2021contemporary} and 3 pre-trained symbolic regression methods. \textbf{Sym.} denotes whether the result is symbolic or not.}
\vspace{2mm}
\begin{tabular}{l|ccc|c}
\toprule
\textbf{Method} & \textbf{R2} $\uparrow$ & \textbf{MSE} $\downarrow$ & \textbf{MAE} $\downarrow$ & \textbf{Sym.} \\ \midrule
\textbf{FFX} & 0.9824 & 4.5e+5 & 3.7e+2 & $\cmark$ \\
\textbf{MLP} & 0.9876 & 3.2e+5 & 3.4e+2 & $\xmark$ \\
\textbf{FEAT} & 0.9964 & 9.2e+4 & 1.7e+2 & $\cmark$ \\
\textbf{DSO} & 0.9968 & 8.2e+4 & 9.2e+1 & $\cmark$ \\
\textbf{Operon} & 0.9988 & 2.8e+4 & 9.8e+1 & $\cmark$ \\ \midrule
\textbf{SymbolicGPT} & 0.5233 & 6.9e+6 & 1.7e+3 & $\cmark$ \\
\textbf{NeSymReS} & \multicolumn{3}{c|}{N/A to $>$3 variables} & $\cmark$ \\
\textbf{T-JSL} & \multicolumn{3}{c|}{N/A to $>$2 variables} & $\cmark$ \\ \midrule
\textbf{Ours} & \textbf{0.9990} & \textbf{1.7e+4} & \textbf{8.6e+1} & $\cmark$ \\
\bottomrule
\end{tabular}
\label{tbl:sr}
\vspace{-5mm}
\end{table}

\subsection{Physical Scientific Discovery}

We consider 6 strong baselines for evaluation: (i) \textbf{Chain-of-Thoughts (CoT)} prompting~\citep{wei2022chain} solves the problem by looking at step-by-step solutions from examples. We provide 5 examples with an explanation to CoT as the initial solution. (ii) \textbf{FunSearch}~\cite{romera2023mathematical} utilizes evolutionary strategy to avoid local optimum. We adopt the given hyperparameters from the original implementation with 2 optimization histories and 4 explorers. We set the number of iterations to 20, yielding the same number of solutions evaluated, for a fair comparison to other methods. (iii) \textbf{Eureka}~\citep{ma2024eureka} generates multiple solutions in each iteration to improve the success rate of the generated code. We keep the hyperparameters from the original implementation. (iv) \textbf{Optimization by PROmpting (OPRO)}~\cite{yang2024large} highlights the advantages of involving a sorted optimization trajectory. We set the hyperparameters to be equal to \textbf{Eureka} except for the number of historical optimization steps. In all these works (i-iv), we notice the temperatures for LLM inference are all 1.0, which is equal to the exploring temperature in our method, so we denote them with 0 exploiter. We also consider 2 variants of our method: (v) \textbf{Ours (no bilevel)} removes the bilevel optimization by only searching with LLM. (vi) \textbf{Ours (no exploit)} removes the exploitation by setting the temperature to 1.0 all the time.

\begin{table}[t]
\centering
\caption{\textbf{Comparison with population-based molecule design}. We compare our method against a traditional population-based molecule design method GhemGE~\cite{yoshikawa2018population} and report the results of molecule design tasks \textbf{(e-h)}.}
\vspace{2mm}
\begin{tabular}{l|cccc}
\toprule
\textbf{Method} & \textbf{(e)} $\downarrow$ & \textbf{(f)} $\downarrow$ & \textbf{(g)} $\downarrow$ & \textbf{(h)} $\downarrow$ \\ \midrule
\textbf{GhemGE} & 4.8e-3 & 1.8 & 1.5 & 9.8e-5 \\
\textbf{Ours} & \textbf{1.3e-4} & \textbf{1.1e-1} & \textbf{5.4e-1} & \textbf{3.6e-5} \\
\bottomrule
\end{tabular}
\label{tbl:molecule}
\vspace{-2mm}
\end{table}

\begin{table}[t]
\centering
\caption{\textbf{Experiment in imaginary constitutive law}. We construct an imaginary constitutive law to keep LLM from cheating by memorization and report the results of our method and baselines.}
\vspace{2mm}
\begin{tabular}{l|cccc}
\toprule
\textbf{Method} & \textbf{FunSearch} & \textbf{Eureka} & \textbf{OPRO} & \textbf{Ours} \\ \midrule
\textbf{Loss} & 105.0 & 89.1 & 98.0 & \textbf{1.3e-3} \\
\bottomrule
\end{tabular}
\label{tbl:imaginary}
\vspace{-5mm}
\end{table}

We present our experiments against the 8 designed tasks and show the results in Table~\ref{tbl:benchmark}. Compared to baselines (i-iv), our method is significantly better by a number of magnitudes. When the bilevel optimization is removed from our method, the performance drops dramatically, but still statistically better than baselines (i-iv), indicating the choice of hyperparameters and the integration of exploitation is helpful for the task. When we remove the exploitation but restore the bilevel optimization, we notice the performance grows back. It has comparable performance compared to our method in \textbf{(d)} or even better results in \textbf{(h)}. However, in some tasks, especially hard ones (e.g., \textbf{(b)} and \textbf{(f)}) that we care more in reality, the performance gap is over $50\%$, indicating the effectiveness of our exploit-and-explore strategy. We also present the loss trend in task \textbf{(a)} in Figure~\ref{fig:benchmark}, our method outstands with a much lower loss and a converging trend.

We also compare our method with traditional methods in each specific area to demonstrate the generalizability of our method. First, we reformulate our constitutive law search task \textbf{(a)} into a symbolic regression task by (i) capture the ground-truth output (the stress tensors) as the supervision, and (ii) separate the 9 output dimension into 9 independent problems and ensemble them for evaluation. Note that these modifications dramatically simplified the original task: we removed back-propagation through time (BPTT) and directly discover the constitutive law without surrogate loss. We evaluate 14 traditional baselines in SRBench~\cite{cava2021contemporary} and 3 data-driven pre-trained baselines. We select the top few baselines in Table~\ref{tbl:sr} and show the rest in the Appendix~\ref{sup:sr}. As shown the table, our method topped on this task even with a much more challenging setting. Also, since our method depends on the in-context learning ability of LLMs, it has little constraint in the number of variables than the data-driven pre-trained baselines. For moledule design tasks, we also compare our method with GhemGE~\cite{yoshikawa2018population}, which employs a population-based molecule design algorithm. As shown in Table~\ref{tbl:molecule}, our method presents a much lower loss, demonstrating the general effectiveness of our method.

\begin{figure}[t]
    \centering
    \includegraphics[width=0.9\linewidth]{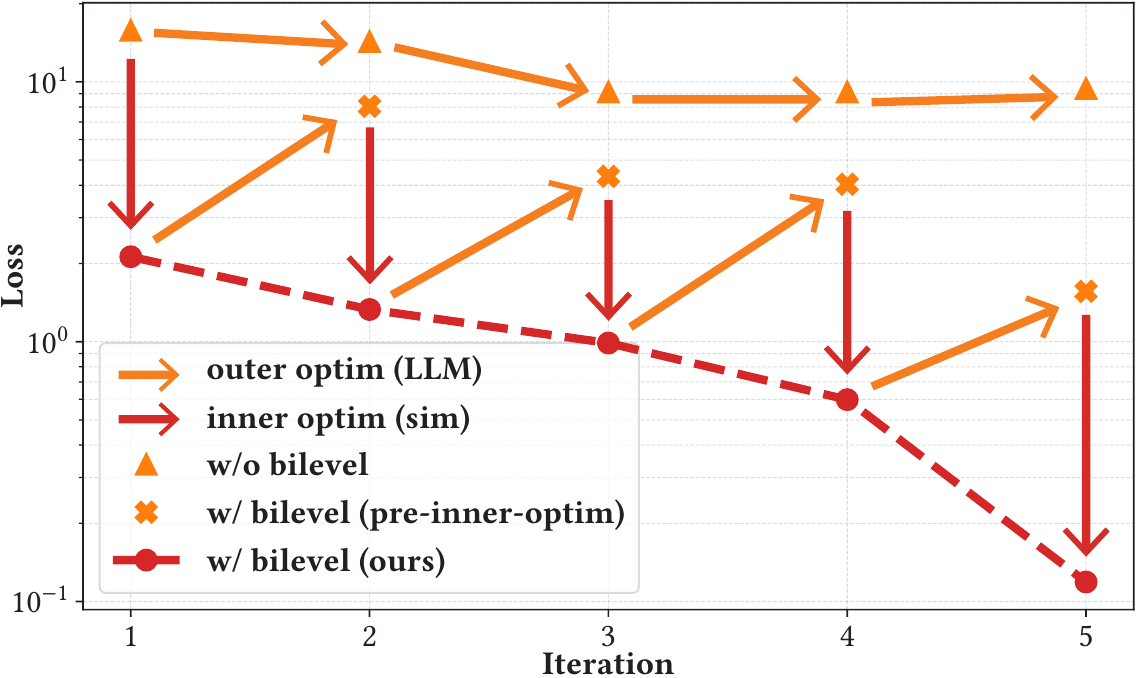}
    \vspace{-2mm}
    \caption{\textbf{Ablation on bilevel optimization.} We denote the optimization trajectory {\color{red}with} and {\color{orange}without} out bilevel optimization with red dot and orange triangle respectively. We visualize the intermediate step of our method {\color{orange}before} the inner-level optimization using orange cross. We also highlight the {\color{orange}outer LLM optimization} and {\color{red}inner simulation optimization} using orange and red arrows.}
    \vspace{-2mm}
    \label{fig:bilevel}
\end{figure}

\subsection{Ablation Study}

\paragraph{Generalization or Memorization}

In order to figure out if the improvement introduced by our method is merely because the LLM saw the solutions in its training phase, we design an experiment ablating it by making it invent an imaginary constitutive law that does not exist on the earth. We mix the constitutive law of von Mises plasticity, granular material, and weakly compressible fluid by 50\%, 30\%, and 20\%, so that the new constitutive law represents an imaginary material whose behavior is extremely complex. We repeat our experiment setup as in Figure 1. We compare our method against the baselines and report the performances in Table~\ref{tbl:imaginary}. As shown in the table, our method can still discover the constitutive law with a low quantitative loss. From our observation, there is \textit{very little visual difference} between the ground-truth material and the optimized constitutive law. We show the discovered constitutive law in Appendix~\ref{sup:imaginary}.

\paragraph{Bilevel Optimization is the Key}

Here we evaluate the importance of bilevel optimization in Figure~\ref{fig:bilevel} using the task \textbf{(h)}. Comparing the blue triangle curve and the red dot curve, which represent the LLM-driven outer-level optimization and the simulation-driven inner-level optimization, it is easy to conclude that the loss performance with bilevel optimization is better. Nevertheless, we are also interested in how bilevel optimization works inside each optimization step and how much LLMs and simulations help respectively. As shown as a zigzag curve, we found that LLMs and simulations help each other over all optimization steps: the next proposal from LLMs will be better with simulation-optimized results, and vice versa. We argue that LLMs and simulations have different expertise: LLMs are generalist scientists who have cross-discipline knowledge, while simulations are domain experts who have specialized knowledge.

\begin{figure}[t]
    \centering
    \includegraphics[width=0.57\linewidth]{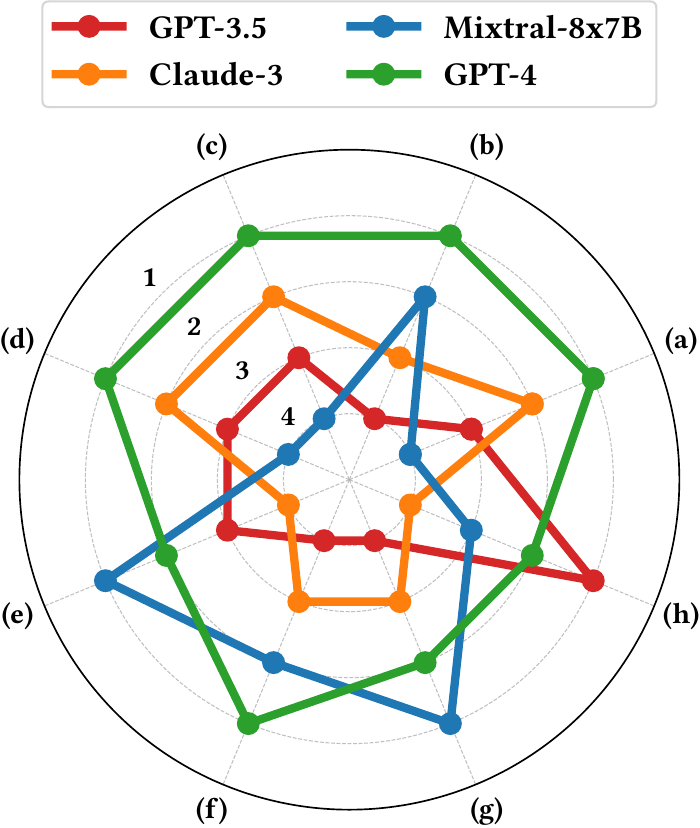}
    \vspace{-2mm}
    \caption{\textbf{Ablation on the backbone LLM.} We compare the performances of 4 selected backbone LLMs and report the rank of them. A outer curve indicates a better performance.}
    \vspace{-3mm}
    \label{fig:llm}
\end{figure}

\paragraph{LLM Backbone} In addition to GPT-4~\cite{openaigpt4blog}, we repeat the experiments in Table~\ref{tbl:benchmark} using 3 additional LLM backbones: (i) GPT-3.5~\cite{ouyang2022training}, (ii) Claude-3-Sonnet~\cite{anthropic2024introducing}, and (iii) Mixtral-8x7B~\cite{jiang2024mixtral}, and report the rank of them in Figure~\ref{fig:llm}. Indicated by the largest area, GPT-4, as our choice, statistically outperforms the other methods. Interestingly, we found Claude-3-Sonnet is the second top method on most of constitutive law search task, while Mixtral-8x7B even tops on 2 molecule design tasks. As a result, our workflow also works for other LLMs, however, our suggestion for practitioners is to try GPT-4 as the first choice but also consider open-source model (e.g., Mixtral-8x7B) for budget or customizability.

\paragraph{Exploitation v.s. Exploration}

\begin{figure*}[t]
    \centering
    \includegraphics[width=\linewidth]{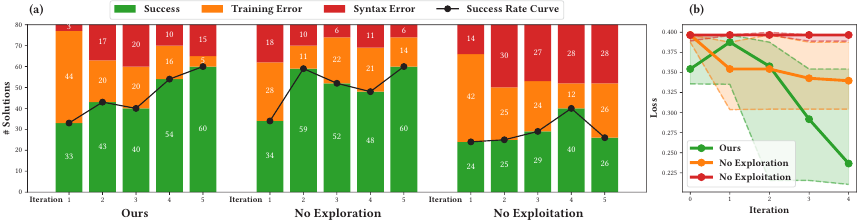}
    \vspace{-7mm}
    \caption{\textbf{Ablation on exploration-exploitation.} (a) Histogram of solutions that are valid for simulation (\eqref{eq:bilevel_2}) across iterations. (b) Loss ($\mathcal{L}$ in \secref{ssec:bilevel}) of the best solution averaged across seeds at different iterations, where the shading indicates the min/max values.}
    \label{fig:ee}
    \vspace{-4mm}
\end{figure*}

We visualize the statistics of the simulation execution status in Figure~\ref{fig:ee} (a) using the task \textbf{(b)}, which is one of the most challenging tasks in our experiments. When the exploitation is removed, the error rate dramatically increases, as shown by a decrease in green bars. It leads to a degeneration in the performance of the methods with exploitation as shown in Figure~\ref{fig:ee} (b). However, even though the success rate remains high, when exploration is removed, the optimization result is still worse than keeping them both. We argue that exploration is significant when the optimization problem is challenging, especially in our case, where the search space is highly non-linear and unstructured and resulting in numerous local optimum.

\subsection{Case Study}

\begin{figure}[t]
    \centering
    \includegraphics[width=0.9\linewidth]{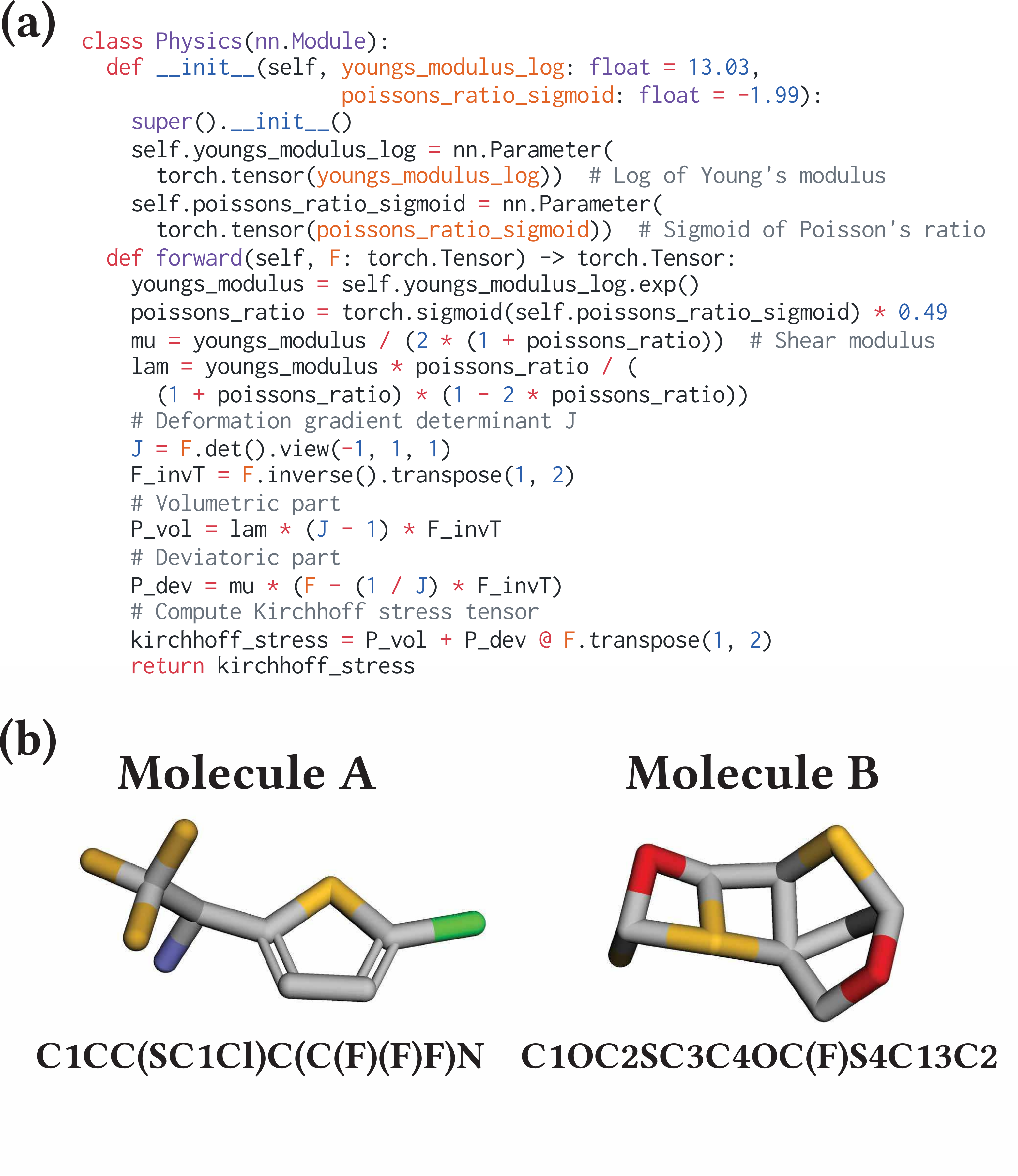}
    \vspace{-10mm}
    \caption{\textbf{Case Study.} \textbf{(a)} We give a concrete example of the searched constitutive law. \textbf{(b)} We provide 2 novel molecules optimized for different objectives with their SMILES stings.}
    \label{fig:case}
    \vspace{-3mm}
\end{figure}

\paragraph{Constitutive Law Search} We provide a trimmed snippet of our searched constitutive law in Figure~\ref{fig:case} (a) for task \textbf{(a)} where a highly non-linear material is provided as the trajectory to fit. We reformat the code slightly to fit into the text, where the complete example can be found in the Appendix. Starting from a linear material, our method is able to automatically generate the constitutive law with a quadratic deviatoric term. Note that our method also provides a concrete implementation of \texttt{\_\_init\_\_} function that defines the continuous parameters in the computational graph for later inner-level optimization.

\paragraph{Molecule Design} When comparing the two molecules with respect to their HOMO-LUMO energy gap based on optimized results from the LLM as shown in Figure~\ref{fig:case} (b), we observe distinct characteristics in each: (i) \textbf{Molecule A} (gap-0) includes sulfur and chlorine atoms attached to a ring, coupled with a trifluoromethyl group, introducing electron-withdrawing effects, and (ii) \textbf{Molecule B} (gap-2) includes oxygen (notably in ethers) and sulfur within the ring structures introducing localized non-bonding electron pairs.
Furthermore, the overall structure of Molecule B is more complex than that of Molecule A, containing multiple rings.
An intriguing aspect of Molecule B, which might initially defy expectations, is the presence of a single fluorine atom. 
The high electronegativity of fluorine typically leads to electron density withdrawal, influencing the gap value. 
However, due to the complexity of Molecule B's structure, the impact of the fluorine atom is somewhat localized, thereby not significantly altering the gap value.

%% file: src/2_related_work.tex
\section{Related Work}
\label{sec:related_work}

\subsection{Automated Scientific Discovery}
Automated scientific discovery, enhanced by machine learning methods, serves as a powerful accelerator for research, enabling scientists to generate hypotheses, design experiments, interpret vast datasets, and unearth insights that may elude traditional scientific methodologies~\citep{ai4science2023impact, kramer2023automated, wang2023scientific}. 
This multifaceted process unfolds through two synergistically linked stages: hypothesis formation and the collection and analysis of experimental data. 
The integration of automated systems not only augments the scientific inquiry process but also streamlines the discovery pipeline, from conceptualization to empirical validation. 
This paper places a particular emphasis on, but is not limited to, constitutive law discovery and molecular design. 
These areas exemplify the profound impact of automation in unraveling complex material behaviors and in the innovative design of molecules with tailored properties.
Automatic identification of constitutive material models has been a long-standing problem and recent works utilizes differentiable simulation~\citep{du2021diffpd,ma2023learning,ma2021risp} to address it as a system identification problem.
Leveraging machine learning and artificial intelligence, researchers are able to predict molecular behavior, optimize chemical structures for specific functions, and thus, rapidly accelerate the development of new drugs, materials, and chemicals~\citep{jin2018junction, zhou2019optimization, schneider2018automating}.

\subsection{Large Language Models and Agents}

The advancement of Large Language Models (LLMs) such as ChatGPT and GPT-4 has sparked considerable interest in their potential as autonomous agents \citep{brown2020language,openaichatgptblog,openaigpt4blog}. Recent developments have shown that LLMs can be enhanced to solve complex problems by creating and utilizing their own tools, as demonstrated in the LATM framework \citep{sumers2024cognitive}, and by acting as optimizers in the absence of gradients, as seen in the OPRO methodology \citep{yang2024large}. These approaches signify a shift towards more independent and versatile LLM-based agents capable of generating solutions through self-crafted tools and optimization techniques \citep{cai2024large,yao2023react,yao2023tree}, showcasing their evolving problem-solving capabilities.
In the realm of scientific discovery, LLMs have begun to make significant contributions, particularly in mathematics and computational problems. The FunSearch method \citep{romera2023mathematical} pairs LLMs with evaluators to exceed known results in extremal combinatorics and online bin packing, illustrating LLMs' ability to discover new solutions to established problems. Similarly, AlphaGeometry's success \citep{trinh2024solving} in solving olympiad-level geometry problems without human demonstrations highlights the potential of LLMs in automating complex reasoning tasks. These examples underline the transformative impact of LLMs in pushing the boundaries of scientific inquiry and automated reasoning.

\subsection{Bilevel Optimization}
Bilevel optimization involves a hierarchical structure with two levels of optimization problems, where the solution to the upper-level problem is contingent upon the outcome of the lower-level problem~\citep{colson2007overview}. 
Bilevel optimization problems are inherently more complex than their single-level counterparts due to the nested nature of the optimization tasks and the intricate interdependencies between them.
Recent advancements have focused on developing efficient algorithms, including evolutionary algorithms~\citep{sinha2017review}, gradient-based approaches~\citep{liu2022general}, and approximation techniques~\citep{sinha2017evolutionary}, to tackle the computational challenges presented by the non-convex and non-differentiable characteristics of many bilevel problems. 
Among a wide span of application domains of bilevel optimization, neural architecture search (NAS)~\citep{liu2018darts, bender2018understanding, cai2018proxylessnas, xue2021rethinking} is prominent and close to the problem setting in this paper: the upper level optimizes the discrete neural network architecture while the lower level optimizes the continuous weights of the neural network. 
However, typical NAS methods require a predefined search space, constraining the exploration of discrete network architectures to manually specified boundaries. 
Our framework distinguishes itself by employing LLM encoded with general knowledge and gets rid of the limitations imposed by manual design constraints.

%% file: src/5_conclusion.tex
\section{Conclusion}
\label{sec:conclusion}

We consider a few limitations and future directions. (i) Although we prompt the LLM to generate pseudo-code plans and comments, it is generally hard to ensure the interpretability of LLM-generated solutions. (ii) Since the LLM-generated codes are executed directly without any filtering in our application, there exists potential AI safety risk that hazards the operating system. (iii) Our method only utilizes the internal knowledge of LLMs as the prior, where in reality people design manual constraints and rule to regularize and improve the optimization~\cite{udrescu2020ai}. We leave these domain-specific applications and human feedback-based regularization methods as our future work. (iv) The performance our method highly depends on the differentiablity of the generated code. However, Zero-order optimizers~\cite{hansen2006cma} should also shine since the number of continuous parameters is relatively limited. (v) LLM inference requires large computational resources and thus increases expense. For example, it spends around \$10 for our method to complete one task using GPT-4, which will be increasingly inacceptable when the number of iteration grows. (vi) Due to the reuse of previously generated solutions in our proposed top-k heap, the KV cache in LLM will be highly similar between neighbor iterations. It opens a gate for recent KV cache optimization methods~\cite{zheng2023efficiently} to speedup our method by KV cache reusing.

In conclution, we present \METHOD{}, a bi-level optimization framework: LLMs serve as knowledgeable and adaptable thinkers, formulating scientific solutions like physics equations or molecule structures; concurrently, simulations operate as platforms for experimentation, offering observational feedback and optimizing continuous components like physical parameters. We focused on two scientific problems: constitutive law search and molecular design. Our approach outperforms other LLM-based benchmark methods, delivering consistent, robust, and nearly monotonic improvement. Furthermore, it shows exceptional ability in identifying unknown, true constitutive laws and molecular structures. Remarkably, our system generates innovative solutions that, despite being unconventional, are deemed reasonable after being thoroughly analyzed by experts in their respective domains. We view our process as a trailblazer, establishing a new paradigm for utilizing LLMs and simulations as bilevel optimization to further advancements in physical scientific discoveries.

%% file: src/6_impact.tex
\section*{Acknowledgements}

We would like to thank Bohan Wang, Ziming Liu, Zhuoran Yang, Liane Makatura, Megan Tjandrasuwita, and Michael Sun for the valuable discussion. The mesh ``Stanford Bunny'' in Figure~\ref{fig:pipeline} is from The Stanford 3D Scanning Repository. This work is supported by MIT-IBM Watson AI Lab.

\section*{Impact Statement}

This paper presents work whose goal is to advance the field of Machine Learning. There are many potential societal consequences of our work, none which we feel must be specifically highlighted here.

%% file: src/n_appendix.tex
\appendix
\onecolumn

\section{Full Prompts}
\label{sec:prompt}

System prompt for constitutive law discovery:

\begin{lstlisting}[style=markdownstyle]
You are an intelligent AI assistant for coding, physical simulation, and scientific discovery.
Follow the user's requirements carefully and make sure you understand them.
Your expertise is strictly limited to physical simulation, material science, mathematics, and coding.
Keep your answers short and to the point.
Do not provide any information that is not requested.
Always document your code as comments to explain the reason behind them.
Use Markdown to format your solution.
You are very familiar with Python and PyTorch.
Do not use any external libraries other than the libraries used in the examples.
\end{lstlisting}

System prompt for molecule design:

\begin{lstlisting}[style=markdownstyle]
You are an intelligent AI assistant for coding, molecule design, and scientific discovery.
Follow the user's requirements carefully and make sure you understand them.
Your expertise is strictly limited to physical simulation, material science, chemistry, molecule design, mathematics, and coding.
Keep your answers short and to the point.
Do not provide any information that is not requested.
Always document your code as comments to explain the reason behind them.
Use Markdown to format your solution.
You are very familiar with PyTorch.
Your are very familiar with the SMILES notation (Simplified Molecular-Input Line-Entry System).
Do not use any external libraries other than the libraries used in the examples.
\end{lstlisting}

Coding format prompt for elastic constitutive law discovery:

\begin{lstlisting}[style=markdownstyle]
## Format Requirements

### PyTorch Tips
1. When element-wise multiplying two matrix, make sure their number of dimensions match before the operation. For example, when multiplying `J` (B,) and `I` (B, 3, 3), you should do `J.view(-1, 1, 1)` before the operation. Similarly, `(J - 1)` should also be reshaped to `(J - 1).view(-1, 1, 1)`. If you are not sure, write down every component in the expression one by one and annotate its dimension in the comment for verification.
2. When computing the trace of a tensor A (B, 3, 3), use `A.diagonal(dim1=1, dim2=2).sum(dim=1).view(-1, 1, 1)`. Avoid using `torch.trace` or `Tensor.trace` since they only support 2D matrix.

### Code Requirements

1. The programming language is always python.
2. Annotate the size of the tensor as comment after each tensor operation. For example, `# (B, 3, 3)`.
3. The only library allowed is PyTorch. Follow the examples provided by the user and check the PyTorch documentation to learn how to use PyTorch.
4. Separate the code into continuous physical parameters that can be tuned with differentiable optimization and the symbolic constitutive law represented by PyTorch code. Define them respectively in the `__init__` function and the `forward` function.
5. The first output of the `forward` function is the updated deformation gradient. Always remember the second output of the `forward` function is Kirchhoff stress tensor, which is defined by the matrix multiplication between the first Piola-Kirchhoff stress tensor and the transpose of the deformation gradient tensor. Formally, `tau = P @ F^T`, where tau is the Kirchhoff stress tensor, P is the first Piola-Kirchhoff stress tensor, and F is the deformation gradient tensor. Do not directly return any other type of stress tensor other than Kirchhoff stress tensor. Compute Kirchhoff stress tensor using the equation: `tau = P @ F^T`.
6. The proposed code should strictly follow the structure and function signatures below:

```python
import torch
import torch.nn as nn

class Physics(nn.Module):

    def __init__(self, param: float = DEFAULT_VALUE):
        """
        Define trainable continuous physical parameters for differentiable optimization.
        Tentatively initialize the parameters with the default values in args.

        Args:
            param (float): the physical meaning of the parameter.
        """
        super().__init__()
        self.param = nn.Parameter(torch.tensor(param))

    def forward(self, F: torch.Tensor) -> torch.Tensor:
        """
        Compute Kirchhoff stress tensor from deformation gradient tensor.

        Args:
            F (torch.Tensor): deformation gradient tensor (B, 3, 3).

        Returns:
            kirchhoff_stress (torch.Tensor): Kirchhoff stress tensor (B, 3, 3).
        """
        return kirchhoff_stress
```

### Solution Requirements

1. Analyze step-by-step what the potential problem is in the previous iterations based on the feedback. Think about why the results from previous constitutive laws mismatched with the ground truth. Do not give advice about how to optimize. Focus on the formulation of the constitutive law. Start this section with "### Analysis". Analyze all iterations individually, and start the subsection for each iteration with "#### Iteration N", where N stands for the index. Remember to analyze every iteration in the history.
2. Think step-by-step what you need to do in this iteration. Think about how to separate your algorithm into a continuous physical parameter part and a symbolic constitutive law part. Describe your plan in pseudo-code, written out in great detail. Remember to update the default values of the trainable physical parameters based on previous optimizations. Start this section with "### Step-by-Step Plan".
3. Output the code in a single code block "```python ... ```" with detailed comments in the code block. Do not add any trailing comments before or after the code block. Start this section with "### Code".
\end{lstlisting}

Coding format prompt for plastic constitutive law discovery:

\begin{lstlisting}[style=markdownstyle]
## Format Requirements

### PyTorch Tips
1. When element-wise multiplying two matrix, make sure their number of dimensions match before the operation. For example, when multiplying `J` (B,) and `I` (B, 3, 3), you should do `J.view(-1, 1, 1)` before the operation. Similarly, `(J - 1)` should also be reshaped to `(J - 1).view(-1, 1, 1)`. If you are not sure, write down every component in the expression one by one and annotate its dimension in the comment for verification.
2. When computing the trace of a tensor A (B, 3, 3), use `A.diagonal(dim1=1, dim2=2).sum(dim=1).view(-1, 1, 1)`. Avoid using `torch.trace` or `Tensor.trace` since they only support 2D matrix.

### Code Requirements

1. The programming language is always python.
2. Annotate the size of the tensor as comment after each tensor operation. For example, `# (B, 3, 3)`.
3. The only library allowed is PyTorch. Follow the examples provided by the user and check the PyTorch documentation to learn how to use PyTorch.
4. Separate the code into continuous physical parameters that can be tuned with differentiable optimization and the symbolic deformation gradient correction model represented by PyTorch code. Define them respectively in the `__init__` function and the `forward` function.
5. The proposed code should strictly follow the structure and function signatures below:

```python
import torch
import torch.nn as nn

class Physics(nn.Module):

    def __init__(self, param: float = DEFAULT_VALUE):
        """
        Define trainable continuous physical parameters for differentiable optimization.
        Tentatively initialize the parameters with the default values in args.

        Args:
            param (float): the physical meaning of the parameter.
        """
        super().__init__()
        self.param = nn.Parameter(torch.tensor(param))

    def forward(self, F: torch.Tensor) -> torch.Tensor:
        """
        Compute corrected deformation gradient from deformation gradient tensor.

        Args:
            F (torch.Tensor): deformation gradient tensor (B, 3, 3).

        Returns:
            F_corrected (torch.Tensor): corrected deformation gradient tensor (B, 3, 3).
        """
        return F_corrected
```

### Solution Requirements

1. Analyze step-by-step what the potential problem is in the previous iterations based on the feedback. Think about why the results from previous constitutive laws mismatched with the ground truth. Do not give advice about how to optimize. Focus on the formulation of the constitutive law. Start this section with "### Analysis". Analyze all iterations individually, and start the subsection for each iteration with "#### Iteration N", where N stands for the index. Remember to analyze every iteration in the history.
2. Think step-by-step what you need to do in this iteration. Think about if the plasticity is needed to improve performance. Remember that plasticity is not necessary. If your analysis supports plasticity, think about how to update deformation gradient using plasticity. Think about how to separate your algorithm into a continuous physical parameter part and a symbolic deformation gradient correction model part. Describe your plan in pseudo-code, written out in great detail. Remember to update the default values of the trainable physical parameters based on previous optimizations. Start this section with "### Step-by-Step Plan".
3. Output the code in a single code block "```python ... ```" with detailed comments in the code block. Do not add any trailing comments before or after the code block. Start this section with "### Code". 
\end{lstlisting}

Coding format prompt for molecule design:

\begin{lstlisting}[style=markdownstyle]
## Format Requirements

### Code Requirements

1. The programming language is always python.
2. Annotate the size of the tensor as comment after each tensor operation. For example, `# (B, 3, 3)`.
3. Separate the code into: (1) python string `SMILES`: the SMILES string describing the molecular topology structure and atomic types, and (2) matrix `coordinates` the 3D coordinates of all atoms. These representations should not include hydrogens.
4. The SMILES string should be valid. Use your knowledge about Simplified Molecular-Input Line-Entry System to help you design a valid one.
5. The number of atoms in the SMILES string should be no less than 8, which means the number of atoms should be >= 8. Try to generate molecule with diverse atoms.
6. The 3D coordinates of the atoms should not be overlapping with each other. In another word, every row in the matrix `coordinates` should be distinct from each other.
7. The `coordinates` matrix is of shape `(N, 3)` where `N` stands for the number of atoms in the molecule. It should be identical to the number of atoms that the proposed SMILES string represents. State out the shape of any matrix defined in the comment as shown in the following example. State out the number of atoms that the SMILES string represents in the comment as shown in the following example.
8. The discrete SMILES string is critical in this problem since it defines the structure and cannot be tuned using differentiable optimization. Please propose different SMILES string from all examples or iterations above to discover and evaluate more structure. This is very important.
9. The proposed code should strictly follow the structure and function signatures below:

```python
SMILES: str # N atoms

coordinates: list[list[float]] # (N, 3)
```

### Solution Requirements

1. Analyze step-by-step what the potential problem is in the previous iterations based on the feedback. Think about why the results from previous molecule structure mismatched with the ground truth. Do not give advice about how to optimize. Focus on the formulation of the SMILES string. Start this section with "### Analysis". Analyze all iterations individually, and start the subsection for each iteration with "#### Iteration N", where N stands for the index. Remember to analyze every iteration in the history.
2. Think step-by-step what you need to do in this iteration. Think about how to separate your algorithm into a continuous 3D coordinate system part and a discrete SMILES string part. Remember the SMILES string proposed should always be different from previous iterations. After propose the new SMILES string, compute and count step-by-step how many atoms it contains. The continuous parameter should follow the number of atoms in the SMILES string. Describe your plan in pseudo-code, written out in great detail. Start this section with "### Step-by-Step Plan".
3. Output the code in a single code block "```python ... ```" with detailed comments in the code block. After the SMILES string, compute the number of atoms in it by counting. Remember that the number of atoms in the SMILES string should be no less than 8, which means the number of atoms should be >= 8. Try to generate molecule with diverse atoms. Do not add any trailing comments before or after the code block. Start this section with "### Code".
\end{lstlisting}

\section{More Explanations}

\subsection{Data Workflow}

The full input to LLM has 3 main parts: (i) system prompt, (ii) iteration information, and (iii) format prompt. For the system prompt, we insert it into the LLM at the beginning or input it as a special instruction depending on the type of LLM. For the iteration information, we first concatenate the code and its feedback and then simply stack the top $K$ solutions. Finally, we append the format prompt at the end of the prompt to regularize the expected output. From our experiments, it is important to keep the order of prompts to ensure the performance and the successful parsing. More precisely, we show this process in the following python-like code:

\begin{lstlisting}[style=pythonstyle]
prompts = []
prompts.append(system_prompt)
for solution in reversed(solutions.topk()):
    iteration_prompt = solution.code + '\n' + solution.feedback
    prompts.append(iteration_prompt)
prompts.append(format_prompt)
full_prompt = '\n'.join(prompts)
\end{lstlisting}

\subsection{Differences to Symbolic Regression Task}

\begin{itemize}
    \item Our problem focuses on loss-guided general scientific discovery, which is a super-set of regular regression problems. In the constitutive law search tasks, we do not directly feed the input/output pair to our method. Instead, we consider a much more challenging task: apply the generated constitutive law recursively and use the overall loss as the performance metric. Concretely, a classic SR methods solve $\arg\min_f\lVert f(X)-y\rVert$ given $<X,y>$ pairs, whereas our method solves $\arg\min_f\lVert g(f(X))\rVert$ given $<X,g(f(X))>$ pairs and $g$ is a complex function like physical simulation. It is easy to construct $g$ to cover the former case using the later formulation, proving the generality of our problem setup. We formulate our problem as such to reflect a more realistic scenario in scientific discovery, where direct supervision is extremely sparse.
    \item Our method supports arbitrary number of input variables and output features, where most of SR methods~\cite{valipour2021symbolicgpt} have limitation on the number of input and output. The input limitation strongly caps the complexity of tasks they can solve, and the output limitation forces them ignore the structural correlation between each output dimension. In a comparison, our method supports arbitrary problem settings thanks to the code-based representation, which enables multi-dimensional arrays and tensor operations.
    \item Our model adapts to multi-discipline application easily, while traditional SR methods typically incorporate with domain-experts' priors via hard-coded constraints and heuristic~\cite{udrescu2020ai}, which is limited, domain-specific, and difficult to customize. Our method is built upon LLMs pre-trained on internet-level data that contains multi-discipline natural languages, mathematical expressions, and codes. As a result, it is easy for users to customize it and adapt to their own diverse applications via natural language guidance.
\end{itemize}

\section{More Experiments}

\subsection{Symbolic Regression}
\label{sup:sr}

We present the full results of the comparison to symbolic regression methods in Table~\ref{tbl:sr-full}

\begin{table}[h]
\centering
\caption{\textbf{Symbolic Regression}}
\begin{tabular}{l|ccc|c}
\toprule
\textbf{Method} & \textbf{R2} $\uparrow$ & \textbf{MSE} $\downarrow$ & \textbf{MAE} $\downarrow$ & \textbf{Symbolic} \\ \midrule
\textbf{AIFeynman}~\cite{udrescu2020ai} & 0.05105 & 22814675.8 & 2520.0 & $\cmark$ \\
\textbf{DSR}~\cite{petersen2020deep} & 0.57527 & 10966411.0 & 2045.0 & $\cmark$ \\
\textbf{BSR}~\cite{jin2019bayesian} & 0.66526 & 8642965.0 & 1938.6 & $\cmark$ \\
\textbf{AdaBoost}~\cite{schapire2003boosting} & 0.75058 & 6439962.9 & 1777.7 & $\xmark$ \\
\textbf{GP-GOMEA}~\cite{virgolin2021improving} & 0.77734 & 5749076.4 & 1580.1 & $\cmark$ \\
\textbf{SBP-GP}~\cite{virgolin2019linear} & 0.81773 & 4706077.0 & 1367.5 & $\cmark$ \\
\textbf{LightGBM}~\cite{ke2017lightgbm} & 0.83368 & 4294433.7 & 1129.9 & $\xmark$ \\
\textbf{XGBoost}~\cite{chen2016xgboost} & 0.87775 & 3156500.5 & 1109.2 & $\xmark$ \\
\textbf{MRGP}~\cite{arnaldo2014multiple} & 0.91074 & 2304682.5 & 950.5 & $\cmark$ \\
\textbf{EPLEX}~\cite{la2019probabilistic} & 0.91851 & 2104070.1 & 122.2 & $\cmark$ \\
\textbf{FFX}~\cite{mcconaghy2011ffx} & 0.93124 & 1775263.7 & 801.7 & $\cmark$ \\
\textbf{MLP} & 0.98240 & 454461.5 & 366.3 & $\xmark$ \\
\textbf{FEAT}~\cite{cava2018learning} & 0.98761 & 319800.6 & 336.1 & $\cmark$ \\
\textbf{DSO}~\cite{mundhenk2021seeding} & 0.99642 & 92374.9 & 168.6 & $\cmark$ \\
\textbf{Operon}~\cite{kommenda2020parameter} & 0.99684 & 81577.9 & 92.4 & $\cmark$ \\ \hline
\textbf{SymbolicGPT}~\cite{valipour2021symbolicgpt} & 0.52333 & 6862154.7 & 1680.7 & $\cmark$ \\
\textbf{NeSymReS}~\cite{biggio2021neural} & \multicolumn{3}{c|}{N/A to $>$3 variables} & $\cmark$ \\
\textbf{T-JSL}~\cite{li2022transformer} & \multicolumn{3}{c|}{N/A to $>$2 variables} & $\cmark$ \\ \midrule
\textbf{Ours} & \textbf{0.99901} & \textbf{17424.6} & \textbf{86.4} & $\cmark$ \\
\bottomrule
\end{tabular}
\label{tbl:sr-full}
\end{table}

\subsection{Longer Iteration}

In order to further investigate the potential of our method and ablate the hyper-parameters for practitioners, we add a new study in terms of the number of iterations (question-answering cycles). We repeat our experiment in Table~\ref{tbl:benchmark} with a prolonged number of iterations to 20 and report the performance in Table~\ref{tbl:longer}.

\begin{table}[h]
\centering
\caption{\textbf{Longer Iteration}}
\begin{tabular}{l|cccc|cccc}
\toprule
\textbf{\#Iterations} & \textbf{(a)} $\downarrow$ & \textbf{(b)} $\downarrow$ & \textbf{(c)} $\downarrow$ & \textbf{(d)} $\downarrow$ & \textbf{(e)} $\downarrow$ & \textbf{(f)} $\downarrow$ & \textbf{(g)} $\downarrow$ & \textbf{(h)} $\downarrow$ \\ \midrule
\textbf{5} & 5.2e-5 & 2.1e-1 & 6.0e-2 & \textbf{1.4e-12} & \textbf{1.3e-4} & 1.1e-1 & 5.4e-1 & 3.6e-5 \\
\textbf{20} & \textbf{4.2e-6} & \textbf{4.0e-4} & \textbf{2.5e-3} & \textbf{1.4e-12} & \textbf{1.3e-4} & \textbf{6.5e-2} & \textbf{1.2e-1} & \textbf{5.6e-6} \\ \midrule
\textbf{Improvement} & +1138.1\% & +52400.0\% & +2300.0\% & 0.0\% & 0.0\% & +69.2\% & +350.0\% & +542.9\% \\ \bottomrule
\end{tabular}
\label{tbl:longer}
\end{table}

As shown in the table, the number of iterations turns out to be a determining hyper-parameter with significant impart on the performance. While it has little affect on relatively easier tasks, it dramatically improves the performance of the most challenging tasks including \textbf{(b)} and \textbf{(c)}. For practitioners, the number of iteration should be first considered as the most important hyper-parameter when adapting our method to their own tasks.

\section{More Results}

\subsection{Constitutive Law Discovery (a)}

The best solution on task \textbf{(a)} optimized by our method:
\begin{lstlisting}[style=pythonstyle]
import torch
import torch.nn as nn

class Physics(nn.Module):
    # Best values from the training curves
    DEFAULT_YOUNGS_MODULUS_LOG = 13.03
    DEFAULT_POISSONS_RATIO_SIGMOID = -1.99

    def __init__(self, youngs_modulus_log: float = DEFAULT_YOUNGS_MODULUS_LOG, poissons_ratio_sigmoid: float = DEFAULT_POISSONS_RATIO_SIGMOID):
        """
        Define trainable continuous physical parameters for differentiable optimization.
        Initialize the parameters with the best values from previous feedback.
        """
        super().__init__()
        # Initialize the parameters as trainable parameters
        self.youngs_modulus_log = nn.Parameter(torch.tensor(youngs_modulus_log))  # Log of Young's modulus
        self.poissons_ratio_sigmoid = nn.Parameter(torch.tensor(poissons_ratio_sigmoid))  # Sigmoid of Poisson's ratio

    def forward(self, F: torch.Tensor) -> torch.Tensor:
        """
        Compute Kirchhoff stress tensor from deformation gradient tensor.

        Args:
            F (torch.Tensor): Deformation gradient tensor (B, 3, 3).

        Returns:
            kirchhoff_stress (torch.Tensor): Kirchhoff stress tensor (B, 3, 3).
        """
        # Convert the parameters to their actual values
        youngs_modulus = self.youngs_modulus_log.exp()  # (1,)
        poissons_ratio = torch.sigmoid(self.poissons_ratio_sigmoid) * 0.49  # (1,)

        # Lame parameters
        mu = youngs_modulus / (2 * (1 + poissons_ratio))  # Shear modulus (1,)
        lam = youngs_modulus * poissons_ratio / ((1 + poissons_ratio) * (1 - 2 * poissons_ratio))  # First Lame parameter (1,)

        # Deformation gradient determinant J and its reshape for operations (B,)
        J = F.det().view(-1, 1, 1)

        # Inverse transpose of F for stress computation (B, 3, 3)
        F_invT = F.inverse().transpose(1, 2)

        # Compute first Piola-Kirchhoff stress tensor P (B, 3, 3)
        # Volumetric part
        P_vol = lam * (J - 1) * F_invT

        # Deviatoric part combining neo-Hookean behavior
        # This accounts for the near incompressible nature of the material
        P_dev = mu * (F - (1 / J) * F_invT)

        # Compute Kirchhoff stress tensor tau by multiplying the first Piola-Kirchhoff with the transpose of F (B, 3, 3)
        kirchhoff_stress = P_vol + P_dev @ F.transpose(1, 2)

        return kirchhoff_stress
\end{lstlisting}

\subsection{Constitutive Law Discovery (b)}

The best solution on task \textbf{(b)} optimized by our method:
\begin{lstlisting}[style=pythonstyle]
import torch
import torch.nn as nn

class Physics(nn.Module):
    
    def __init__(self, gamma: float = -0.07):  # Based on best value from iteration 5
        """
        Initialize gamma as a trainable parameter which will be used for scaling the soft 
        deformation correction.
        
        Args:
            gamma (float): scaling factor for the deformation correction.
        """
        super().__init__()
        self.gamma = nn.Parameter(torch.tensor(gamma))  # Initialize gamma, (1,)

    def forward(self, F: torch.Tensor) -> torch.Tensor:
        """
        Compute corrected deformation gradient tensor F, by applying a soft correction
        proportional to the deviation of its determinant from 1, effectively guiding the
        gradient towards physically realistic states.

        Args:
            F (torch.Tensor): deformation gradient tensor (B, 3, 3).

        Returns:
            F_corrected (torch.Tensor): corrected deformation gradient tensor (B, 3, 3).
        """
        # Compute determinant of F and create a condition based on its value, (B,)
        J = torch.det(F)  # (B,)
        
        # Apply a smooth step function as a deviation condition, (B,)
        J_deviation_condition = torch.tanh(J - 1)  # (B,)

        # Prepare for correction, taking into account the batch dimension (B,)
        gamma_correction = self.gamma * J_deviation_condition.view(-1, 1, 1)  # (B, 1, 1)

        # Identity matrix, expanded for batch size (B, 3, 3)
        I = torch.eye(3, device=F.device).repeat(F.size(0), 1, 1)  # (B, 3, 3)

        # Correct F by pulling towards identity matrix when determinant deviates from 1, (B, 3, 3)
        F_corrected = F - gamma_correction * (F - I)  # (B, 3, 3)

        return F_corrected
\end{lstlisting}

\subsection{Constitutive Law Discovery (c)}

The best solution on task \textbf{(c)} optimized by our method:
\begin{lstlisting}[style=pythonstyle]
import torch
import torch.nn as nn

# The default value for elastic_limit is set to the best from the last iteration, and
# we initialize a new parameter for capturing the hardening effect
DEFAULT_ELASTIC_LIMIT = 0.92
DEFAULT_HARDENING_FACTOR = 0.1

class Physics(nn.Module):
    
    def __init__(self, elastic_limit: float = DEFAULT_ELASTIC_LIMIT, 
                 hardening_factor: float = DEFAULT_HARDENING_FACTOR):
        """
        Define trainable continuous physical parameters for differentiable optimization.
        
        Args:
            elastic_limit (float): the parameter determining the initial yield strength.
            hardening_factor (float): the parameter controlling the rate of hardening.
        """
        super().__init__()
        self.elastic_limit = nn.Parameter(torch.tensor(elastic_limit))  # ()
        self.hardening_factor = nn.Parameter(torch.tensor(hardening_factor))  # ()

    def forward(self, F: torch.Tensor) -> torch.Tensor:
        """
        Compute corrected deformation gradient from deformation gradient tensor.

        Args:
            F (torch.Tensor): deformation gradient tensor (B, 3, 3).

        Returns:
            F_corrected (torch.Tensor): corrected deformation gradient tensor (B, 3, 3).
        """
        # Obtain the polar decomposed rotation (R) and stretch (S)
        U, S, V = torch.svd(F)  # U: (B, 3, 3), S: (B, 3), V: (B, 3, 3)
        R = U @ V.transpose(-2, -1)  # R: (B, 3, 3)

        # Correct the S tensor with hardening
        # Assuming hardening affects the elastic limit linearly with accumulated plastic strain
        plastic_strain = torch.relu(S - self.elastic_limit)  # Presumed plastic strain
        hardening_adjustment = 1.0 + (self.hardening_factor * plastic_strain)
        
        S_clamped = torch.min(S, self.elastic_limit * hardening_adjustment)  # Clamp S with hardening
        S_corrected = torch.diag_embed(S_clamped)  # S_corrected: (B, 3, 3)
        
        F_corrected = R @ S_corrected  # (B, 3, 3) Corrected deformation gradient tensor
        
        # Ensure volume preservation
        J = torch.det(F).view(-1, 1, 1)  # (B, 1, 1) Determinant of the input F for volume
        J_corrected = torch.det(F_corrected).view(-1, 1, 1)  # (B, 1, 1) Determinant of the corrected F
        volume_ratio = (J / J_corrected) ** (1/3)
        F_corrected = F_corrected * volume_ratio  # (B, 3, 3) Volume-preserved F_corrected
        
        return F_corrected
\end{lstlisting}

\subsection{Constitutive Law Discovery (d)}

The best solution on task \textbf{(d)} optimized by our method:
\begin{lstlisting}[style=pythonstyle]
import torch
import torch.nn as nn

class Physics(nn.Module):

    DEFAULT_VALUE = 0.0  # Best guess based on previous behavior

    def __init__(self, param: float = DEFAULT_VALUE):
        """
        Define trainable continuous physical parameters for differentiable optimization.
        The parameter modulates corrections towards nearly isochoric behavior.
        """
        super().__init__()
        self.param = nn.Parameter(torch.tensor([param]))  # Scalar modulation parameter

    def forward(self, F: torch.Tensor) -> torch.Tensor:
        """
        Symbolic deformation gradient correction model.
        """
        # Compute the determinant of the deformation gradient (volumetric change)
        J = torch.det(F)  # (B,)

        # Compute the volumetric part of the deformation gradient: J^(1/3)*I
        I = torch.eye(3).to(F.device)  # (3, 3)
        # Expand the identity matrix to the entire batch
        I = I.view(1, 3, 3).expand(F.size(0), -1, -1)  # (B, 3, 3)
        # Calculate volumetric part
        vol_deform = torch.pow(J, 1.0 / 3.0).view(-1, 1, 1) * I  # (B, 3, 3)

        # Calculate the deviatoric part of F: divide F by J^(1/3)
        dev_deform = F / torch.pow(J, 1.0 / 3.0).view(-1, 1, 1)  # (B, 3, 3)

        # Modulate correction by self.param and construct the correction term
        correction = self.param * (I - dev_deform)  # (B, 3, 3)

        # Combine the volumetric part with the deviatoric correction
        F_corrected = vol_deform + correction  # (B, 3, 3)

        return F_corrected  # (B, 3, 3)
\end{lstlisting}

\subsection{Molecule Design (e)}

The Top-20 solution on task \textbf{(e)} optimized by our method:
\begin{enumerate}
\item \texttt{C1=CC=C(Br)C=C1C2=CN=CC=C2}
\item \texttt{C1=CC=C(I)C=C1C2=CC=NC=C2}
\item \texttt{C1=CN(C=C1)C2=CC(=CC(=C2)F)O}
\item \texttt{C1C2CC3CC(C1)C(C2)(C3)N}
\item \texttt{C1CC2OC1COC2=O}
\item \texttt{C1=CC=C(C=C1)C2=NC(Cl)=NC=C2}
\item \texttt{C1=CC=C(I)C=C1C2=NC(C(F)(F)F)=CN=C2}
\item \texttt{C1N2C3C4OC(C5)C13C24C5}
\item \texttt{C1=CC=C2N=CC=C(Br)C2=C1}
\item \texttt{C=CC1C(=O)NC(=S)N1C}
\item \texttt{C1=CC=CS1C2=NC=CC(Cl)=C2}
\item \texttt{C1OC2C(O)C3C(N)C1C23}
\item \texttt{O=C(NC1=CC=CC=C1)C2CCOCC2Cl}
\item \texttt{C1=CC=C2C(=C1)N=CN2C3=CC=CC=C3Br}
\item \texttt{C1OC2C3N4C5C6C7C8C1C2C3C4C5C6C7C8}
\item \texttt{C1=CC=C(C=C1)C2=CN=CC=C2C=CCl}
\item \texttt{C1=CC=C(S1)C2=CC=C(O2)C(F)F}
\item \texttt{C1=CC=C2C(=C1)C(=NN2)C3=CC=CC=C3Br}
\item \texttt{C1=CC=C2N=C(C(=O)NC2=C1)C3=CC=CS3}
\item \texttt{C1=CC=C(C=C1)C2=NN=C(S2)Cl}
\end{enumerate}

\subsection{Molecule Design (f)}

The Top-20 solution on task \textbf{(f)} optimized by our method:
\begin{enumerate}
\item \texttt{SC(F)(F)CC(Cl)(Cl)N}
\item \texttt{C1=CSC(=C1)NNC(=O)CF}
\item \texttt{C1C2C3CSC1N2OC3F}
\item \texttt{C1CSC(Cl)C(N)C1Cl}
\item \texttt{CC(C(=O)O)NC(F)(F)F}
\item \texttt{O=C(O)C(F)=C(Cl)C(Br)C=O}
\item \texttt{C1CSC(C(=O)O)N=C1F}
\item \texttt{C1=CC(=CS1)C2=CC=C(Br)C(F)=C2}
\item \texttt{O=S(=O)(N)C1=CC=C(Br)C=C1}
\item \texttt{CC(C(=O)O)C(Cl)C(Br)C(N)C(I)}
\item \texttt{C1SC(Cl)C(C1)C(=O)O}
\item \texttt{FC1=CC=C(C=C1)CS}
\item \texttt{C1CSC2C3CC(Cl)C1C23}
\item \texttt{C1=CC(=O)N(C2=CS1)C2=O}
\item \texttt{C1=CC=C(N)C(Cl)=C1Cl}
\item \texttt{C1SCC2C1C1=C(C=O)C=CC1C2Cl}
\item \texttt{C1=CC2=C(N=C(I)C=C2)C=C1}
\item \texttt{NC(CS)C(C(=O)O)Cl}
\item \texttt{C1=CC=NC2=C1C(=O)SC2=CCBr}
\item \texttt{C1=CC=C2C(=C1)C(=CS2)Cl}
\end{enumerate}

\subsection{Molecule Design (g)}

The Top-20 solution on task \textbf{(g)} optimized by our method:
\begin{enumerate}
\item \texttt{c1cc(sc1Cl)C(C(F)(F)F)N}
\item \texttt{C(C(Cl)Cl)C1=CSC(N)=N1}
\item \texttt{C1C2CC3NC1C3C2O}
\item \texttt{C1OC2C3NOC1C3C2}
\item \texttt{C1C(Cl)C2CNC1C2O}
\item \texttt{C1(=CC(=C(N1Cl)O)C\#N)S}
\item \texttt{C1COC2C3CC4(NC2C34)O1}
\item \texttt{c1cc(c(c(c1)Cl)O)C(F)(F)F}
\item \texttt{O1CCN2CC(F)C2C1}
\item \texttt{N1CC2OC(F)C2C1}
\item \texttt{C1(O)C2C(NC2C1C)C}
\item \texttt{C1=CC2=C(S1)C(=CC(=C2)P)I}
\item \texttt{C1CC2NCC(C1)C2O}
\item \texttt{C1CC2SCC(C1)N2}
\item \texttt{C1=NC2=CS(=O)(=O)N=C12}
\item \texttt{C1OC2C3CC(S)C1C23}
\item \texttt{C1C2C(NC=O)C(Cl)C1OC2}
\item \texttt{O=S(=O)(c1ccccc1)N}
\item \texttt{C1C=CC(O1)(F)N2C=CC(Br)=C2Cl}
\item \texttt{O=N(=O)C=C(C)C=C(C)N=O}
\end{enumerate}

\subsection{Molecule Design (h)}

The Top-20 solution on task \textbf{(h)} optimized by our method:
\begin{enumerate}
\item \texttt{CC(NC(=O)C(Cl)C(=O)O)CSC}
\item \texttt{C1=CC(=CC=C1)C2=NSN=C2}
\item \texttt{FC(F)Oc1ccccc1N}
\item \texttt{O=C1NC(=O)SC2=C1C=CC=C2}
\item \texttt{C1NOC2C1SC1C2N1Cl}
\item \texttt{C1CSCC(N)C1N=C(O)C2=CC=CS2}
\item \texttt{C1C2CC(NC1=O)C(Cl)C2I}
\item \texttt{C1OC2SC3C4OC(F)S4C13C2}
\item \texttt{SC(Cl)(Cl)C1=CC=CC=C1O}
\item \texttt{C1=NOC(=C1)C2C(C(=O)NC2=O)Cl}
\item \texttt{C1OC2C3NCC4S1C23C4}
\item \texttt{C1C2C(N(C1Cl)C2=O)F}
\item \texttt{OC1C2C(O1)N=CS(=O)C2}
\item \texttt{C1OC2C3N4C1SC23N4}
\item \texttt{C1CSC(C2=NC=CS2)N1}
\item \texttt{C1CSCC(N1)C2=CC=CS2}
\item \texttt{O=P1(OP(=O)(OC1=O)C2=CN=CC=C2)I}
\item \texttt{C1CC2(CNC2)C(O1)C3=CN=CN3}
\item \texttt{O=C1SCCN(C1)C2=CN=CO2}
\item \texttt{C1=CC=NC2=C1C(=O)N=C(S2)Cl}
\end{enumerate}

\subsection{Imaginary Constitutive Law}
\label{sup:imaginary}

The best solution on the imaginary constitutive law invention task optimized by our method:
\begin{lstlisting}[style=pythonstyle]
import torch
import torch.nn as nn

# Default values for the physical parameters based on previous iterations
DEFAULT_KAPPA = 0.08
DEFAULT_MU = 0.28

class Physics(nn.Module):
    def __init__(self, kappa: float = DEFAULT_KAPPA, mu: float = DEFAULT_MU):
        """
        Initialize the continuous physical parameters kappa and mu for differentiable optimization.
        """
        super().__init__()
        self.kappa = nn.Parameter(torch.tensor(kappa))  # Bulk modulus correction factor (scalar)
        self.mu = nn.Parameter(torch.tensor(mu))       # Shear modulus correction factor (scalar)

    def forward(self, F: torch.Tensor) -> torch.Tensor:
        """
        Compute the corrected deformation gradient from the deformation gradient tensor.
        """
        B = F.size(0)  # Batch size (scalar)
        I = torch.eye(3, device=F.device).unsqueeze(0).expand(B, -1, -1)  # Identity matrix (B, 3, 3)
        J = torch.det(F).view(-1, 1, 1)  # Jacobian determinant (B, 1, 1)
        
        # Volume correction factor (B, 1, 1)
        vol_correction_factor = torch.clamp(self.kappa * (J - 1), min=0.0, max=1.0)
        vol_correction = vol_correction_factor * I  # Volume correction term (B, 3, 3)
        
        # Compute trace of F for shape correction (B, 1, 1)
        trace_F = F.diagonal(dim1=1, dim2=2).sum(dim=1).view(-1, 1, 1)
        dev_F = F - (trace_F / 3) * I  # Deviatoric part of F (B, 3, 3)
        
        # Shape correction factor (scalar)
        shape_correction_factor = torch.clamp(self.mu, min=0.0, max=1.0)
        shape_correction = shape_correction_factor * dev_F  # Shape correction term (B, 3, 3)
        
        F_corrected = F - vol_correction - shape_correction  # Corrected deformation gradient (B, 3, 3)
        return F_corrected
\end{lstlisting}